\documentclass{article}

\usepackage[nonatbib, preprint]{neurips_2020}

\usepackage[utf8]{inputenc}
\usepackage[T1]{fontenc}
\usepackage[english]{babel}
\usepackage{hyperref}
\usepackage{url}
\usepackage{booktabs}
\usepackage{amsmath}
\usepackage{amsfonts}
\usepackage{amssymb}
\usepackage{amsthm}
\usepackage{nicefrac}
\usepackage{microtype}
\usepackage{graphicx}
\usepackage{epstopdf}
\usepackage{textcomp}
\usepackage{gensymb}
\usepackage{acronym}
\usepackage[linesnumbered,ruled,vlined]{algorithm2e}

\title{Consensus Driven Learning}

\author{%
  Kyle L.~Crandall\\
  Center for High Assurance Computer Systems\\
  U.S. Naval Research Laboratory\\
  Washington, DC 20375 \\
  \texttt{kyle.crandall@nrl.navy.mil} \\
  \And
  Dustin~Webb \\
  AI Influx, LLC \\
  Salt Lake City, Utah 84109 \\
  \texttt{dustin@aiinflux.com} \\
}

\acrodef{CL}{Consensus Learning}
\acrodef{FL}{Federated Learning}
\acrodef{RL}{Reinforcement Learning}
\acrodef{DAC}{Distributed Averaging Consensus}
\acrodef{ML}{Machine Learning}
\acrodef{FIFO Buffer}{First In, First Out Buffer}
\acrodef{A3C}{Asynchronous Advantage Actor Critic}
\acrodef{SGD}{Stochastic Gradient Descent}
\acrodef{SVM}{Support Vector Machine}
\acrodef{DNN}{Deep Neural Network}
\acrodef{CNN}{Convolutional Neural Network}
\acrodef{MSE}{Mean Squared Error}
\acrodef{ReLU}{Rectified Linear Unit}
\acrodef{MLP}{Multi-Layered Perceptron}

\begin{document}

\maketitle

\begin{abstract}
As the complexity of our neural network models grow,  so too do the data and computation requirements for successful training.  One proposed solution to this problem is training on a distributed network of computational devices, thus distributing the computational and data storage loads.  This strategy has already seen some adoption by the likes of Google and other companies.  In this paper we propose a new method of distributed, decentralized learning that allows a network of computation nodes to coordinate their training using asynchronous updates over an unreliable network while only having access to a local dataset.  This is achieved by taking inspiration from Distributed Averaging Consensus algorithms to coordinate the various nodes.  Sharing the internal model instead of the training data allows the original raw data to remain with the computation node.  The asynchronous nature and lack of centralized coordination allows this paradigm to function with limited communication requirements.  We demonstrate our method on the MNIST, Fashion MNIST, and CIFAR10 datasets. We show that our coordination method allows models to be learned on highly biased datasets, and in the presence of intermittent communication failure.
\end{abstract}

\section{Introduction}
With the increase in popularity of \ac{ML}, and as it is applied to more and more applications, the models being trained to solve these problems grow more and more complex and require more and more data to train on, and more and more computational power to process this data through the complex models.  This growth, coupled with growing concerns about data security have made distributed learning methods more appealing as they allow for an array of compute nodes to work on the same problem simultaneously, reinforcing the learning done by other nodes.  This creates an environment with significantly more scaleability compared to fully centralized methods.  An increase in the number of compute nodes allows for an increase in the amount of data being processed at once, as the each node would maintain its own local dataset.  Some early work in distributed learning introduced methods such as HOGWILD! \cite{Recht2011} and DOGWILD! \cite{Noel2014}.  These methods involve many different compute nodes training a model, occasionally overwriting the weights in a central, master model, and occasionally overwriting their own weights with the ones on the central model in the case of HOGWILD!, or overwriting and copying each other as is the case in DOGWILD!.  As Google began to explore this problem, they developed a series of asyncronous learning techniques, including their \ac{A3C} method \cite{Mnih2016}.  In these methods, compute nodes accumulate updates, then periodically push them to a global model, then copy that model and begin accumulating more updates.  In 2019, they proposed an overarching framework for this sort of approach they refer to as \ac{FL}.  In this framework, models are sent to remote compute nodes, they are trained on local data on those nodes, then the updates are sent back to the master node, which aggregates the data from the various compute nodes \cite{Bonawitz2019}.  These methods provide advantages in scalability, and only rely on local data by sharing the current iteration of the model in training rather than the raw data.  One of the major attractions of such methods is that they adress contemporary concerns about data security \cite{Bonawitz2019}.  

While these methods are great for the large scale \ac{ML} applications that are common these days, we are particularly interested in their use with mobile teams of robots.  With the exception of DOGWILD!, all the above methods rely on a central parameter server that tracks the current ``truth'' of the model.  Maintaining a central model allows for good testing and verification of the model during training.  This is undesirable in the case of a mobile robotic team, as it makes the individual robots dependent on a communication channel to the central server to get the latest version of the complete model, which puts severe restrictions on where the robots can move, especially if communication infrastructure does not exists, say in a disaster zone after an earthquake or hurricane.  In addition, these methods, especially the only decentralized model, DOGWIlD!, also require robust, high bandwidth communication, which, is not always in field robotics.

In this paper, we propose an algorithm we call \iac{CL} algorithm.  While the previously discussed algorithms, perform well in industrial and laboratory conditions where high fidelity communication between compute nodes is a reasonable assumption, \ac{CL} is designed to work with minimal communication requirements.  This algorithm is inspired by \ac{DAC} algorithms, a class of algorithms run over a network of nodes, each with a value.  The average of these values is calculated in a distributed manner by sharing individual estimates of the average repeatedly until they converge to a consensus.  Much of the earlier work in \ac{DAC} algorithms focused on averaging constant values.  Of particular interest to us were methods using asynchronous updates \cite{Mehyar2007}. More recent work has focused on time varying signals in continuous time \cite{Chen2012}\cite{George2019}, but much of this work requires synchronized, continuous updates.  Using similar methods, our algorithm reaches a consensus on the model being trained without any centralized coordination.  We implement our algorithm without the extreme bandwidth requirements of methods like DOGWILD!, and even show that it is robust to communications failures.

In this paper, we give an overview of our proposed method, and outline its theoretical advantages and limitations. We then present an example case using this method to train a model on the MNIST dataset.  We perform several variations to the training setup to demonstrate the strengths and limitations of our method.  We show that allowing the nodes to communicate allows them to achieve better performance overall than when they do not, and that this method is robust to various communication failures.  We also show that this method allows the model to learn well, even when local datasets are extremely biased.  We then repeat all of these experiments on the Fashion MNIST and CIFAR10 datasets, and compare the results from all three sets of experiments.

\section{Consensus Learning Algorithms}
\ac{CL} algorithms run on a network of compute nodes with some ability to communicate with each other.  Each compute node uses a standard training method with local data to train, and a consensus algorithm to coordinate the training across the full network.

Our proposed algorithm is inspired by the \ac{DAC} algorithm presented in \cite{Mehyar2007}.  This algorithm has a node send a copy of its estimate to another node, the receiving node takes a weighted average of its estimate and the received estimate, then sends the deltas back to the originating node.  The originating node then subtracts the deltas from its estimate completing the update.  The results in the two estimates moving towards each other, and the mean between the two values in preserved.  If a node is currently in the middle of an update, that node rejects any additional update requests.  Our algorithm is based on this method, however rather than ignoring incoming updates while busy, they are stored in a buffer, and local updates to the values are performed using traditional training methodology.

Our algorithm assumes \iac{ML} problem set up as follows:
\begin{itemize}
	\item a model $y = \phi\left(x \vert p\right)$ where $y$ are the labels, $x$ are the inputs, and $p$ are the parameters.
	\item a loss function $\lambda = L\left(x, y\right)$
	\item a training function $p_{k+1} = \texttt{trainOnBatch}\left(p_k, x, y\right)$
	\item a training dataset with a function that returns a batch of data to train on $x,y = \texttt{getBatch}\left(\right)$
	\item the parameters $p$ are initialized to some value $p_{i0}$.  This value can be different for each node.
\end{itemize}
We also require a communications structure to facilitate the sending of messages between nodes.  We assume this structure has the following capabilities for a given node $i$:
\begin{itemize}
	\item $C_{it}$, a set of all nodes to whom a communication channel exists at time $t$
	\item $U_i$, an input \ac{FIFO Buffer}
	\item $\texttt{sendTo}\left(m,j\right)$, a function that will place message $m$ into $U_j$ if $j \in C_{it}$.
	\item $m, j = \texttt{recvNext}\left(U_i\right)$ a function to retrieve the next message in $U_i$ and its source node, or returns $\emptyset$ if $U_i = \emptyset$.
\end{itemize}

<<<<<<< HEAD
Because our algorithm uses a similar strategy as the \ac{DAC} algorithm in \cite{Mehyar2007} to update the weights across pairs of nodes, there are two types of messages being sent between nodes.  To signal the type of data a message contains, it is prepended with either a \texttt{WEIGHTS} or \texttt{DELTAS} flag.
=======
Because the consensus algorithm uses two different types of messages, each message has a flag at the beginning to specify the type.  Specifically a \texttt{STATE} flag to signal the message contains the current estimate of the averages, and an \texttt{UPDATE} flag that indicates the message contains an update to be applied to an internal state.
>>>>>>> dustin

Our \ac{CL} algorithm is broken into two distinct phases, the local learning phase where the nodes internal model is updated based on local training data, the asynchronous update phase where the nodes update each other to reach towards a consensus on the model.  The local learning phase will perform localized training on $N_i$ batches of data, then move on the the asynchronous update phase.  This phase will start by sending the node's current weights to $M_i$ random nodes in $C_{it}$.  Then the node will process all the messages in its communication buffer $U_i$, then it will return to the local learning phase.  Each node will alternate through these different phases until the learning task is deemed complete.  Algorithm \ref{alg:ConsensusLearning} outlines this process in pseudo code.

\IncMargin{2em}
\begin{algorithm}
	\caption{\ac{CL} Algorithm for Node $i$}
	\label{alg:ConsensusLearning}
	\SetAlgoLined
	\SetKwFunction{getBatch}{getBatch}
	\SetKwFunction{trainOnBatch}{trainOnBatch}
	\SetKwFunction{randSelect}{randSelect}
	\SetKwFunction{sendTo}{sendTo}
	\SetKwFunction{recvNext}{recvNext}
	\SetKwData{packet}{packet}
	$p_i \leftarrow p_{i0}$\;
	\While{running}{
		\tcp{Local Learning phase}
		\For{$j \leftarrow 0$ \KwTo $N_i$}{
			$x \leftarrow$ \getBatch{}\;
			$p_i \leftarrow$ \trainOnBatch{$x,p_i$}\;
		}
		\tcp{Asynchronous Update phase}
		\For{$j \leftarrow 0$ \KwTo $M_i$}{
			\packet $\leftarrow \left(\texttt{WEIGHTS}, p_i\right)$\;
			$c \leftarrow$ \randSelect{$C_{it}$}\;
			\sendTo{\packet,$c$}\;
		}
		\While{$U_i \neq \emptyset$}{
			\packet,$c \leftarrow$ \recvNext{$U$}\;
			\uIf{\packet$\left[0\right] == \texttt{WEIGHTS}$}{
				$p_j \leftarrow$ \packet$\left[1\right]$\;
				$\Delta p \leftarrow \gamma_i\left(p_j - p_i\right)$\;
				\packet $\leftarrow \left(\texttt{DELTAS}, \Delta p\right)$\;
				\sendTo{\packet,$c$}\;
				$p_i \leftarrow p_i + \Delta p$\;
			}
			\ElseIf{\packet$\left[0\right] == \texttt{DELTAS}$}{
				$\Delta p \leftarrow$ \packet$\left[1\right]$\;
				$p_i \leftarrow p_i - \Delta p$\;
			}
		}
	}
\end{algorithm}
\DecMargin{2em}

\section{Experimental Results}
We performed several experiments to demonstrate the utility of this algorithm, and to explore its strengths and limitations, by training \iac{DNN} on the MNIST dataset as it is a common toy case that demonstrates how our method works quite nicely.  In the first set of experiments, the number of updates that are sent out each iteration are varied.  In the second set, we trained the \ac{DNN} on locally biased datasets, and varied the level of bias.  In the last set of experiments, we explored a case where communication is unreliable between nodes by dropping packets at a varying rate.  In all sets of experiments, we compare the results to a control where the \ac{DNN} is trained on a the complete dataset using traditional, non-decentralized methods, we refer to this case as the monolithic model.  Because MNIST is a widely considered to be a toy problem, we also performed these same tests on the Fashion MNIST dataset \cite{Xiao2017} and the CIFAR10 dataset \cite{Krizhevsky2010} to show that it works when confronted with more complex problems as well.

For the MNIST and Fashion MNIST datasets, we trained a \ac{DNN} that is \iac{MLP}.  This \ac{DNN} had a single hidden layer with 72 neurons using \ac{ReLU} activation, and an output layer with 10 neurons using a SoftMax activation.  For the CIFAR10 dataset, we trained \iac{CNN} with two convolutional layers identifying 16 and 32 features with 3x3 pixel windows, \ac{ReLU} activation, and max pooling after each layer (This network architecture was derrived from the tensorflow \cite{KerasTutorial2020}).  The network was then flatened, and a 128 neuron dense layer was applied with \ac{ReLU} activation and dropout.  Finaly a 10 neuron dense layer was used as the output layer with SoftMax activation.  In all the experiments, the \ac{DNN} was trained using a Sparse Categorical Cross-Entropy cost function along with an Adam Optimizer \cite{Kingma2014}.  Each compute node also has a copy of the same validation and test sets so the results are comparable.

\subsection{Data Sharing}
In this experiment, the \ac{DNN} described above was trained on the MNIST dataset distributed over 8 nodes, with each node having a local dataset that is 1/8th of the full dataset.  We vary the number of times a copy of the weights are sent out each cycle ($M_i$).  We do a control test where $M_i=0$, and compare it to cases where $M_i=1,2,$ and $5$.  We also compare these results to the monolithic model results.

\begin{figure}
	\centering
	\begin{tabular}{c c}
		\includegraphics[width=0.45\textwidth]{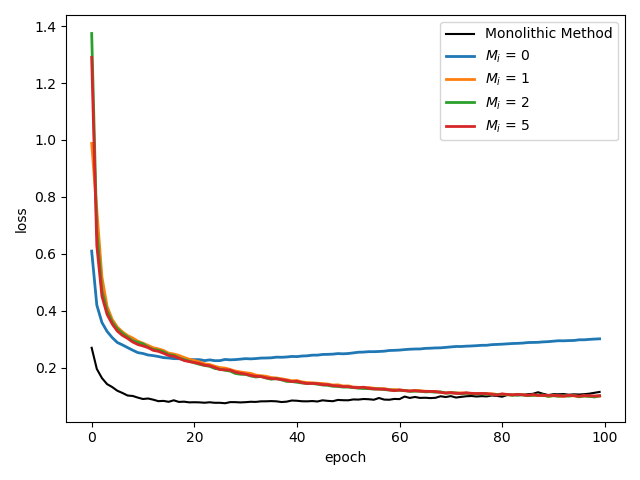} &
		\includegraphics[width=0.45\textwidth]{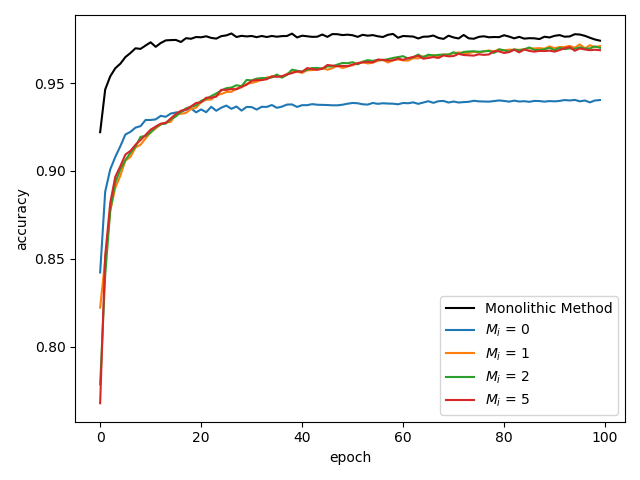} \\
		(a) & (b)
	\end{tabular}
	\caption{Learning performance during first set of experiments on the MNIST Dataset where $M_i$ was varied.  Metrics were evaluated on a validation set.  Subfigure (a) shows the loss, and (b) shows the accuracy.  The metrics were averaged over all the nodes.}
	\label{fig:metrics_update}
\end{figure}

Figure \ref{fig:metrics_update} shows the metrics taken during these tests.  In the case where $M_i=0$, the nodes seemed to learn on their local datasets for a short time, then quickly begin to memorize this dataset (as indicated by the upward trend in the validation loss signal that begins around epoch 20).  When we examine the actual weights, we find that each node learns a distinct model (which is expected given randomized weight initialization).  In this case, these models were not able to achieve the level of performance as the monolithic model.

When we allow the nodes to update each other, the models are able to achieve similar performance to the monolithic model.  However, it took significantly longer to to achieve this performance.  In addition, when examining the weights of the final model, we found that the nodes all learned more or less the same model, as we expect from employing the \ac{DAC} algorithm.  This indicates that the sharing of the model in our proposed algorithm allows relevant information in the local dataset to make its way to the other nodes resulting in a model that performs more like it was trained on the global dataset.

The last key observation of this experiment is that when $M_i \neq 0$, the value of $M_i$ did not seem to dramatically effect the performance of the algorithm.  This means that we can limit the number of updates to reduce the amount of network traffic necessary to run this algorithm.  Though on unreliable networks, multiple updates might by necessary to increase the probability of an update making it to its destination, as discussed later.

\subsection{Training on Locally Biased Data}
To explore the ability of this method to effectively share information between nodes, we considered a case where nodes are given very biased local data while the global dataset remains unbiased.  To create these biased data sets, we first sorted the MNIST Dataset by label.  Given a mix rate $0 \leq r_m \leq 100$, we created 10 datasets where for dataset $i=0...9$, we took $r_m$\% of the data points labeled $i$ and distributed them among the other 9 datasets, leaving $\left(100-r_m\right)\%$ of the data with label $i$ to be added to dataset $i$.  This resulted in 10 datasets with a variable level of bias regulated by $r_m$.  We setup 10 nodes to train on these datasets using the proposed \ac{CL} algorithm, and ran this test for mix rates at intervals of 10\% from 10\% to 90\%.

\begin{figure}
	\centering
	\begin{tabular}{c c}
		\includegraphics[width=0.45\textwidth]{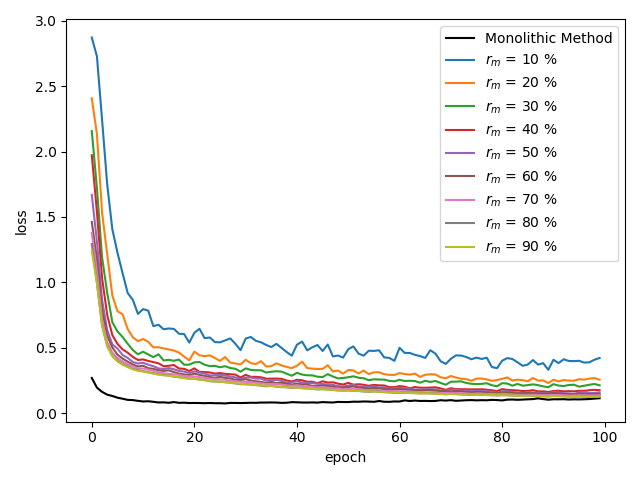} &
		\includegraphics[width=0.45\textwidth]{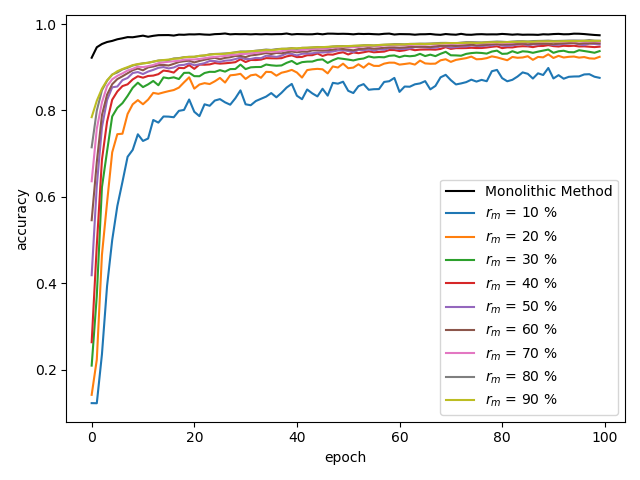} \\
		(a) & (b)
	\end{tabular}
	\caption{Learning performance during second set of experiments on the MNIST Dataset where the \ac{DNN} is trained on locally biased datasets.  Metrics were evaluated on a validation set.  Subfigure (a) shows the loss, and (b) shows the accuracy.  The metrics were averaged over all the nodes.}
	\label{fig:metrics_mix}
\end{figure}

Figure \ref{fig:metrics_mix} shows the results of these tests.  Not shown in this figure are the results when using a mix rate of 0\%, in this case, the learning failed, giving an accuracy of around 10\%, which is what we would expect of a model trained on a dataset consisting of only one label.  However, with a very low mixing rate of even 10\%, the nodes were able to learn to recognize inputs with labels not prominent in their own local set. This does come at a price however, the learning on highly biased datasets takes longer to converge than better mixed datasets.

\subsection{Unreliable Communication}
In this last experiment, we considered cases where communication is not reliable, and packets are dropped.  As was shown in the first set of experiments we presented, increasing $M_i$ does not seem to improve the performance of the algorithm under ideal conditions.  However, when we consider the possibility of the dropped packets, increasing $M_i$ could be used to increase the probability that a packet will make it through.  What is more concerning to us is if this algorithm is still valid if the first packet makes it to its destination, but the reply packet gets dropped.  We ran the same scenario that we ran as part of the first set of experiments, but instead of $M_i$, we vary the probability a reply will get dropped.  We use dropout rates of 0\%, 25\%, 50\%, 75\%, and 100\%.

\begin{figure}
	\centering
	\begin{tabular}{c c}
		\includegraphics[width=0.45\textwidth]{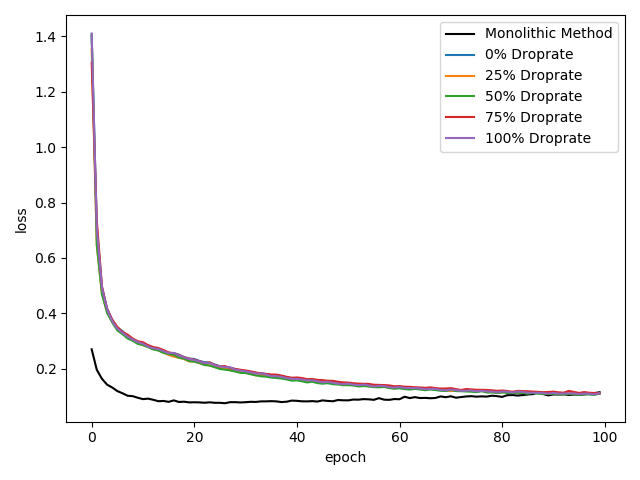} &
		\includegraphics[width=0.45\textwidth]{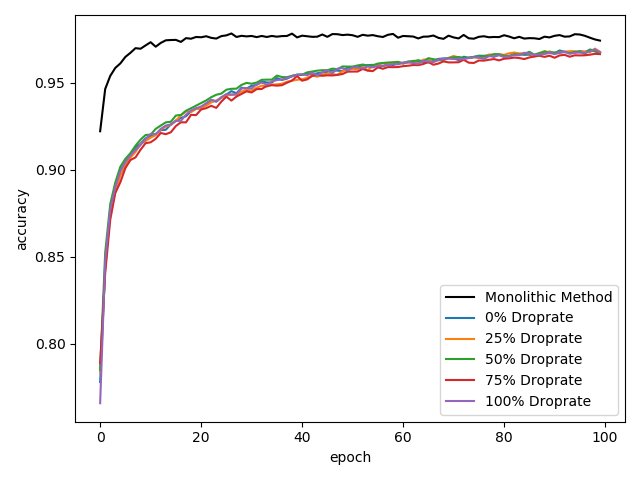} \\
		(a) & (b)
	\end{tabular}
	\caption{Learning performance during last set of experiments on the MNIST Dataset where reply packets are dropped.  Metrics were evaluated on a validation set.  Subfigure (a) shows the loss, and (b) shows the accuracy.  The metrics were averaged over all the nodes.}
	\label{fig:metrics_step}
\end{figure}

The results of these experiments are displayed in Figure \ref{fig:metrics_step}.  As we can see, dropping packets did not seem to greatly effect the performance of the algorithm.  This combined with the results from the first experiment show that this algorithm is fairly robust to intermittent communications failures.  In fact, sending the reply message does not seem to be necessary, at least for these experiments.  However we chose to keep it as an optional part of our algorithm for the sake of the \ac{DAC} algorithm that was the inspiration for this algorithm.

\subsection{Additional Experiments}
\begin{table}
	\caption{Training results on multiple datasets \label{tab:results}}
	\centering
	\begin{tabular}{l r c c c c c c}
		\hline \hline
		&& \multicolumn{2}{c}{MNIST} & \multicolumn{2}{c}{Fashion MNIST} & \multicolumn{2}{c}{CIFAR10} \\
		\textsc{Experiment} && \textsc{Loss} & \textsc{Acc.} & \textsc{Loss} & \textsc{Acc.} & \textsc{Loss} & \textsc{Acc.} \\
		\hline
		Monolithic && 0.0905 & 0.9758 & 0.3528 & 0.8832 & 0.9415 & 0.6822 \\
		\hline
		Update Rate & $M_i=$ \\
		& 0 & 0.2775 & 0.9425 & 0.6376 & 0.8416 & 3.3870 & 0.5395 \\
		& 1 & 0.0924 & 0.9720 & 0.3743 & 0.8685 & 1.1968 & 0.6411 \\
		& 2 & 0.0925 & 0.9727 & 0.3683 & 0.8687 & 1.0734 & 0.6492 \\
		& 5 & 0.0935 & 0.9728 & 0.3698 & 0.8706 & 1.0350 & 0.6416 \\
		\hline
		Mixed Data & $r_m=$ \\
		& 10 & 0.3860 & 0.8853 & 0.8047 & 0.7571 & 2.4799 & 0.2925 \\
		& 20 & 0.2316 & 0.9298 & 0.5967 & 0.8008 & 1.9104 & 0.3852 \\
		& 30 & 0.1936 & 0.9411 & 0.5122 & 0.8243 & 1.6351 & 0.4768 \\
		& 40 & 0.1589 & 0.9526 & 0.4700 & 0.8386 & 1.3562 & 0.5549 \\
		& 50 & 0.1428 & 0.9563 & 0.4393 & 0.8465 & 1.3635 & 0.5567 \\
		& 60 & 0.1340 & 0.9597 & 0.4172 & 0.8537 & 1.1509 & 0.6080 \\
		& 70 & 0.1222 & 0.9637 & 0.4139 & 0.8547 & 1.1921 & 0.6095 \\
		& 80 & 0.1161 & 0.9644 & 0.3948 & 0.8602 & 1.1631 & 0.6252 \\
		& 90 & 0.1127 & 0.9670 & 0.3863 & 0.8640 & 1.2044 & 0.6280 \\
		\hline
		Droprate & $dr=$ \\
		& 0\%   & 0.1012 & 0.9703 & 0.3735 & 0.8685 & 1.0633 & 0.6517 \\
		& 25\%  & 0.0987 & 0.9703 & 0.3662 & 0.8711 & 1.0213 & 0.6537 \\
		& 50\%  & 0.1027 & 0.9697 & 0.3719 & 0.8691 & 0.9947 & 0.6692 \\
		& 75\%  & 0.1027 & 0.9699 & 0.3707 & 0.8688 & 1.0194 & 0.6575 \\
		& 100\% & 0.1057 & 0.9691 & 0.3812 & 0.8648 & 1.0960 & 0.6387 \\	
		\hline \hline	
	\end{tabular}
\end{table}

We also performed the previously mentioned experiments on two additional datasets, the Fashion MNIST Dataset, and the CIFAR10 dataset.  Table \ref{tab:results} shows the metrics for the results of these experiments as well as the MNIST results.  For each dataset, the resulting neural networks were run on a test dataset consisting of data points that the networks did not see during training or validation, and the same test set was used for each experiment.  The resulting metrics are averaged over all the nodes that were used during training. We can see the same patterns in the results with these new datasets that were apparent in the MNIST experiments.  In all cases, When $M_i = 0$, the the decentralized consensus method trained networks that perform poorly compared to when $M_i > 0$, where the trained networks were able to achieve comparable results to the network produced by the monolithic method given enough time (100 epochs in all cases).  Further, increasing $M_i$, and adjusting the dropout rate of return packets did not seem to have a noticeable effect on the convergence and performance of the algorithm in any of our test cases.  Finally, when the data was divided into highly biased subsets, the decentralized consensus method was still able to produce a network that is able to recognize categories not prevalent in a given node's local dataset.  The overall performance increased with the mixing rate in all cases.

\section{Conclusion}
In this paper we propose a Consensus Learning paradigm and an algorithm that implements it.  This algorithm takes inspiration from a Distributed Averaging Consensus method to reach a consensus on the model being trained, while using traditional training algorithms to train the model on local data.  This algorithm does so without significant requirements on the communication network the nodes are operating on, and is robust to intermittent failure in that network.  We demonstrate this algorithm by using it to train a Deep Neural Network model on the MNIST Dataset of handwritten numerals, as well as the Fashion MNIST dataset, and the CIFAR10 dataset.  We demonstrate the robustness of this algorithm to failures in the communication network, as well as showing its capacity to overcome bias in local datasets in all three cases.

\section*{Broader Impacts}
This work does not present any foreseeable societal consequence.

\begin{ack}
The work on this project was funded by the Karles Fellowship at Naval Research Laboratory under number 300000145350
\end{ack}

\bibliographystyle{IEEEtran}
\bibliography{ConsensusLearning}

\end{document}